\documentclass{svmult}

\usepackage{makeidx}         
\usepackage{graphicx}        
\usepackage{multicol}        
\usepackage[bottom]{footmisc}

\usepackage{amsmath}
\usepackage{amsfonts}
\usepackage{booktabs}
\usepackage{subfigure}
\usepackage{verbatim}
\usepackage{url}

\makeindex             

\begin{document}

\title*{Camera Pose Estimation from Sequence of Calibrated Images}
\author{Jacek Komorowski\inst{1}\and
Przemyslaw Rokita\inst{2}}
\institute{
Maria Curie-Sklodowska University,
Institute of Computer Science,
Lublin, Poland
\texttt{jacek.komorowski@gmail.com}
\and Warsaw University of Technology,
Institute of Computer Science,
Warsaw, Poland
\texttt{p.rokita@ii.pw.edu.pl}
}
%
%
\maketitle

\begin{abstract}
In this paper a method for camera pose estimation from a sequence of images is presented. 
The method assumes camera is calibrated (intrinsic parameters are known) which allows to decrease a number of required pairs of corresponding points compared to uncalibrated case. 
Our algorithm can be used as a first stage in a structure from motion stereo reconstruction system.
\end{abstract}


\section{Introduction}
\label{jk:sec:1}

\begin{figure}
\centering
\includegraphics[height=4.7 cm]{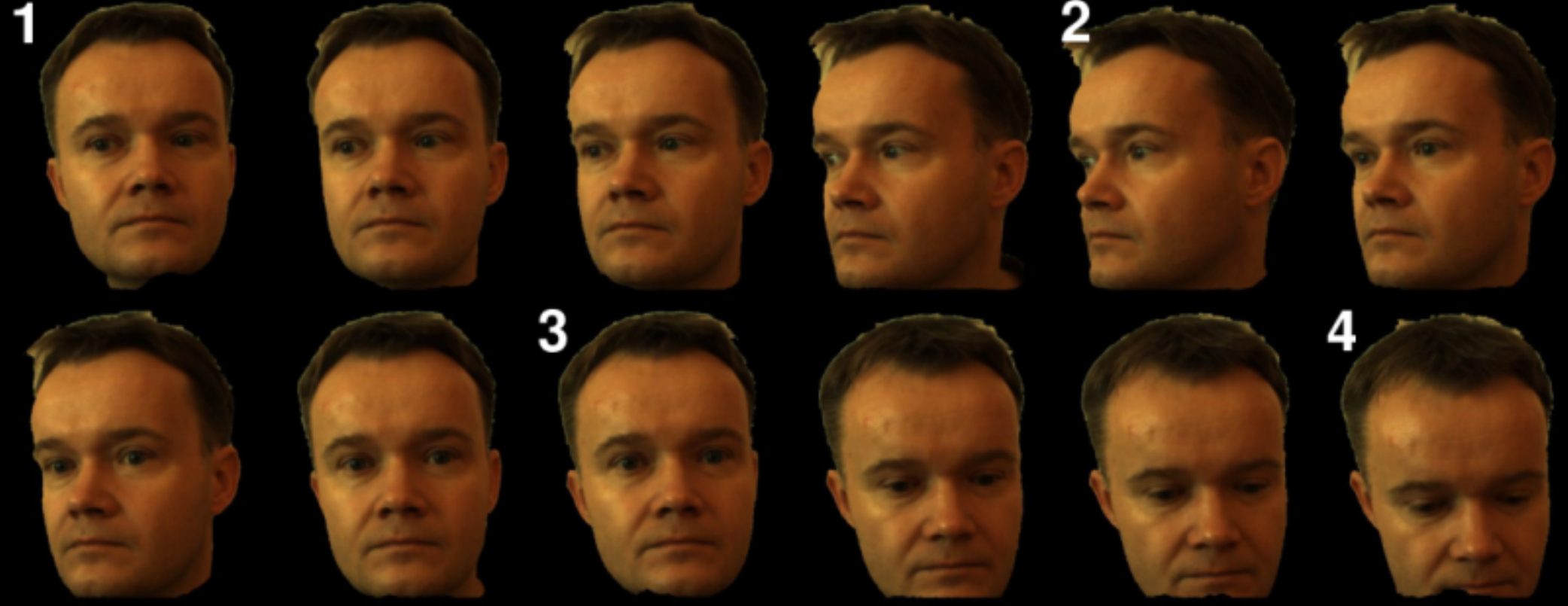}
\caption[]{Examplary images from an input sequence (every 8th image of 120 frames input sequence).
Head is initially positioned in front of the camera (1) and rotates right app. 45$^{\circ}$(2), back to the frontal pose (3) and downward app. 20$^{\circ}$ (4).}
\label{jk:fig:10}

\end{figure}

Motivation for development of the method described in this paper was our prior research on human face reconstruction from a sequence of images from a monocular camera (such as depicted on Fig. \ref{jk:fig:10}).
Such sequence representing an object moving or rotating in front of a fixed camera can be alternatively thought of as a sequence of images of a static object taken by a moving camera and such perspective is adopted in this paper.
Classical multi-view stereo reconstruction algorithms (as surveyed in \cite{jk:sei06}) assume fully calibrated setup, where both intrinsic and extrinsic camera parameters are known for each frame. 
Such algorithms cannot be used in our scenario where an object moves or rotates freely in front of the camera. Even if camera intrinsic parameters are known and fixed during the entire sequence, camera extrinsic parameters (rotation matrix and translation vector relating camera reference frame with the world reference frame) are not known.
So before a multi-view stereo reconstruction algorithm can be used a prior step to estimate camera pose (extrinsic parameters) for each image in the sequence is required.

Such methods usually work by finding a correspondence between feature points on subsequent images and then recovering camera pose and scene structure using matched features.
Human skin has relatively a homogeneous texture and initial experiments showed that methods based on tracking feature points on subsequent images, such as Kanade-Lucas-Tomasi tracker were not performing as expected.
In this paper we describe an alternative approach based on ideas used in modern structure from motion products such as Bundler \cite{jk:snav07} or Microsoft PhotoSynth.
These solutions work by finding geometric relationship (encoded by fundamental matrix) between 2 images of the scene taken from different viewpoints. 
This is usually done by running a robust parameter estimation method (e.g. RANSAC) combined with 7-point or 8-point fundamental matrix estimation algorithm using putative pairs of corresponding features from 2 images.

However, in contrast to aforementioned solutions, in our method we make an assumption about fixed and known (from a prior calibration stage) camera intrinsic parameters. 
When camera intrinsic parameters are known, less pairs of corresponding points are required to recover 2-view scene geometry. This should significantly decrease number of iterations needed by RANSAC to estimate model parameters with a given confidence.

Unfortunately currently known algorithms for estimation of relative pose between 2 calibrated cameras from 5 pairs of corresponding points are very complex and implementations are not freely available.
E.g. Nister 5-point algorithm \cite{jk:nis04} requires SVD, partial Gauss-Jordan elimination with pivoting of a system of of polynomial equations of the third degree and finally finding roots of a 10th degree polynomial. Such complexity can potentially lead to significant numerical errors and make such methods inapplicable in practice.

So aim of our work was twofold: first to design a solution for estimation of extrinsic parameters for a sequence of images from a calibrated camera, and second, to verify that C++ implementation of Nister 5-point algorithm is numerically stable.

\section{Pose estimation method details}
\label{jk:sec:2}

An input to our pose estimation method is a sequence of images from a calibrated, monocular camera, such as depicted on Fig. \ref{jk:fig:10}. The the following steps are done:

\begin{enumerate}
	\item Initial processing: object segmentation from the background and geometric distortions removal. Further processing is done on undistorted and segmented images.
	\item Detection of SIFT features on all images in the sequence.
	\item Estimation of the relative pose between 2 initial images in the sequence:
	
	\begin{enumerate}
		\item Finding pairs of putative matches between SIFT features on both images.
		\item Computation of essential matrix $E_{12}$ relating two images using RANSAC \cite{jk:fis81} with Nister \cite{jk:nis04} solution to 5-point relative pose problem. The relative pose (translation vector $T_{2}$ and rotation matrix $R_{2}$) is recovered from $E_{12}$ as described in \cite{jk:nis04}.
		\item Construction of an initial 3D model (as a sparse set of 3D points) by metric triangulation of pairs of consistent features (reprojection error and distance between feature descriptors are below thresholds) from two images. 
		\item 3D points and camera pose refinement using bundle adjustment method \cite{jk:tri99} to minimize reprojection error.
	\end{enumerate}
	
	\item Iterative estimation of an absolute pose of each subsequent image $I_n$ with respect to 3D model built so far:
	\begin{enumerate}
		\item Finding putative matches between features on the image $I_n$ and 3D points already in the model.
		\item Computation of an absolute pose (translation vector $T_{k}$ and rotation matrix $R_{k}$) of the image $I_k$ with respect to the 3D model. This is done using RANSAC \cite{jk:fis81} with Finsterwalder 3-point perspective pose estimation algorithm \cite{jk:har94}.
		\item Guided matching of features from currently processed image $I_k$ and images processed in the previous steps. New 3D points are generated and added to 3D model (and support of existing 3D points is extended)  by metric triangulation of matching features.
		\item 3D points and camera pose refinement using bundle adjustment method \cite{jk:tri99} to minimize reprojection error.
	\end{enumerate}
	
\end{enumerate}

Final results are depicted on Fig. \ref{jk:fig:90}, where green dots represent recovered camera poses for each image from an input sequence from Fig. \ref{jk:fig:10}.

Additional details on each algorithm step:

\paragraph{Step 2} 
SIFT features \cite{jk:lowe99} are a common choice in modern structure from motion solutions. This is dictated by their invariance to scaling, rotation and, to some extent, lighting variance and small affine image transformations. These properties are important when finding corresponding features on images taken from different viewpoints. At this step SIFT features are found and feature descriptors (represented as vectors from $\mathbb{R}^{128}$) are computed for each image in the sequence. 

\begin{figure}
\centering
\subfigure[]{
\includegraphics[height=2.2 cm]{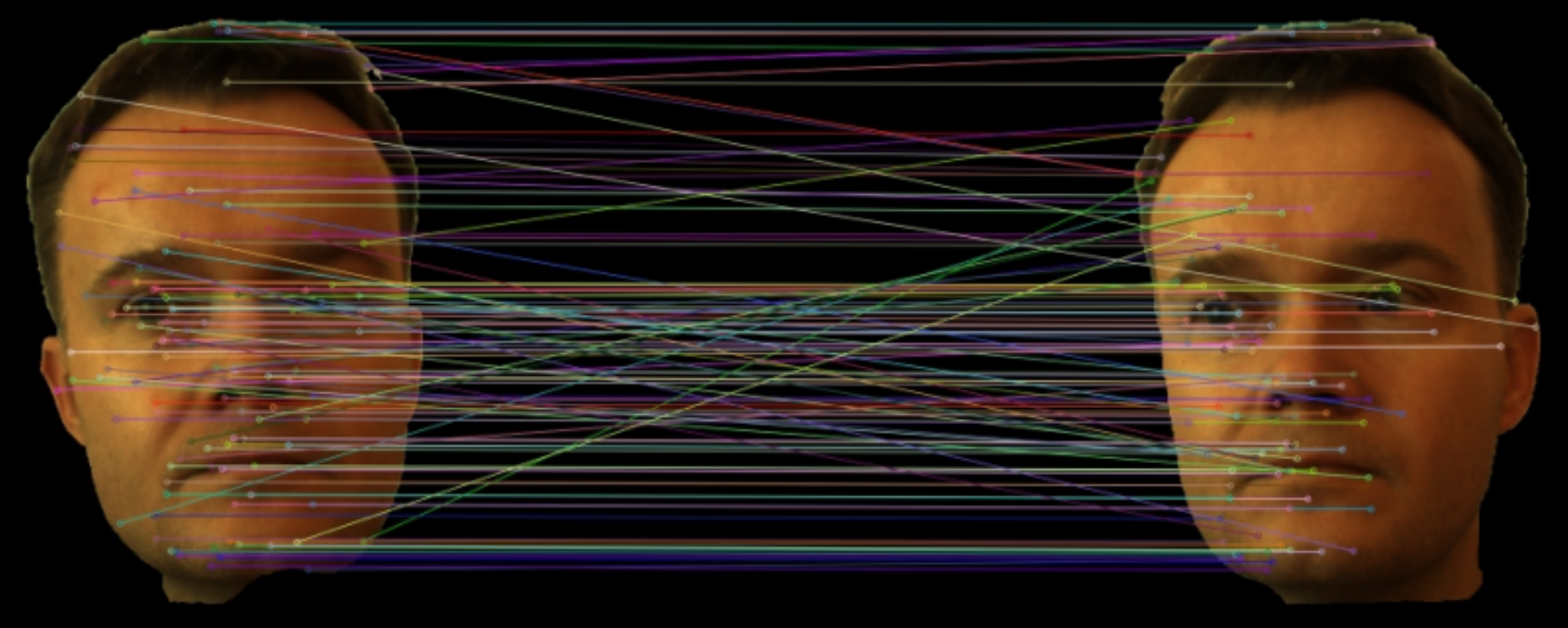}
\label{jk:fig:51}
}
\subfigure[]{
\includegraphics[height=2.2 cm]{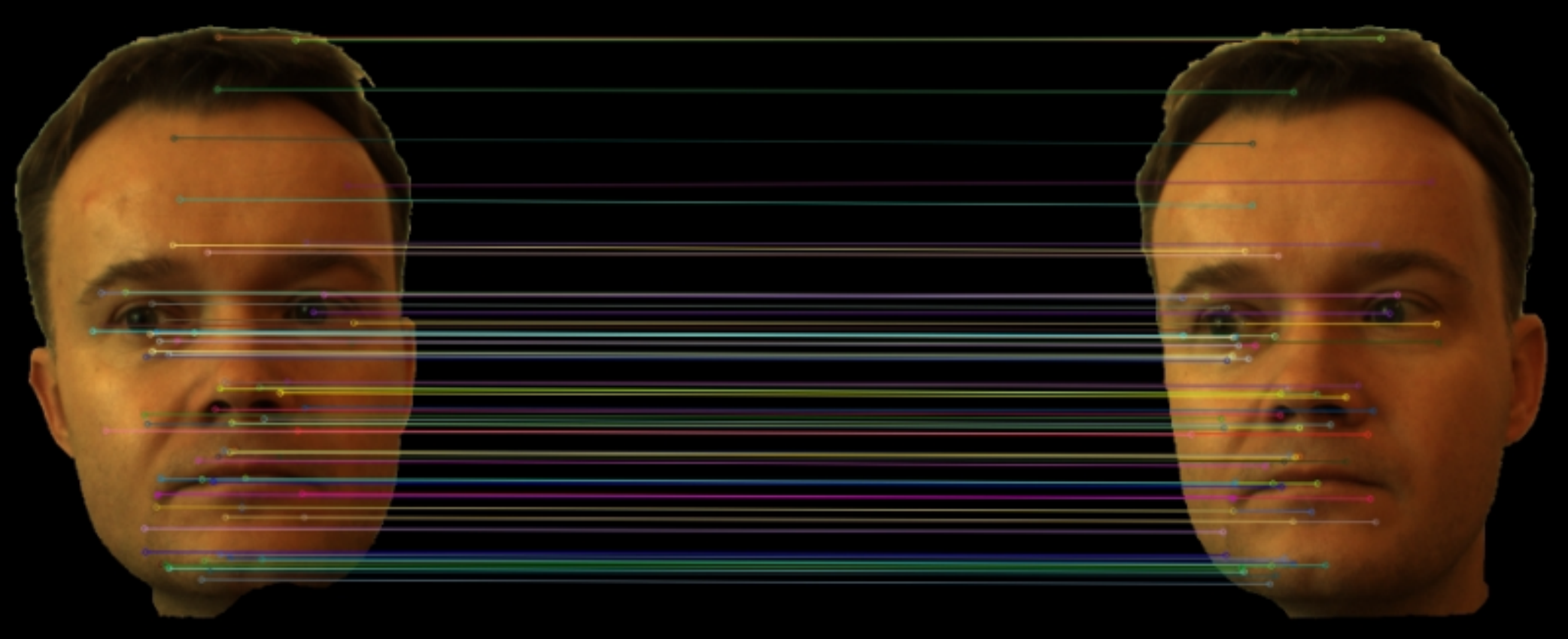}
\label{jk:fig:52}
}

\label{jk:fig:50}
\caption{Pairs of matches between 2 images \subref{jk:fig:51} putative matches \subref{jk:fig:52} matches consistent with epipolar geometry encoded by estimated essential matrix $E$} 
\end{figure}

\paragraph{Step 3a} For each keypoint from the first image the closest (in the feature descriptor space) keypoint from the second image is found. Only pairs fulfilling nearest neighbour ratio criterion (that is ratio of a distance to the corresponding keypoint to the distance to the second-closest keypoint on the other image is below given threshold $\Theta = 1.25$) are kept as putative matches. See Fig. \ref{jk:fig:51}.

\paragraph{Step 3b} RANSAC \cite{jk:fis81} robust parameter estimation is used with our implementation of Nister 5-point algorithm \cite{jk:nis04} to estimate relative pose between 2 cameras from a set of putative point correspondences. Results of this step are: essential matrix $E_{12}$ describing stereo geometry between 2 images from a calibrated camera, rotation matrix $R_2$ and translation vector $T_2$ describing the relative pose of the second image with respect to the first image, consensus set consisting of pairs of matching features consistent with epipolar geometry (see Fig. \ref{jk:fig:52})

\section{Experimental results}

Quantitative evaluation of accuracy of the presented method is difficult due to lack of reliable ground truth data.
In this paper we only present quantitative examination, using a synthetic data, of one key component of our solution, that is estimation of the relative pose of two calibrated cameras using our implementation of Nister 5-point algorithm.


\begin{figure}
\centering
\includegraphics[height=5 cm]{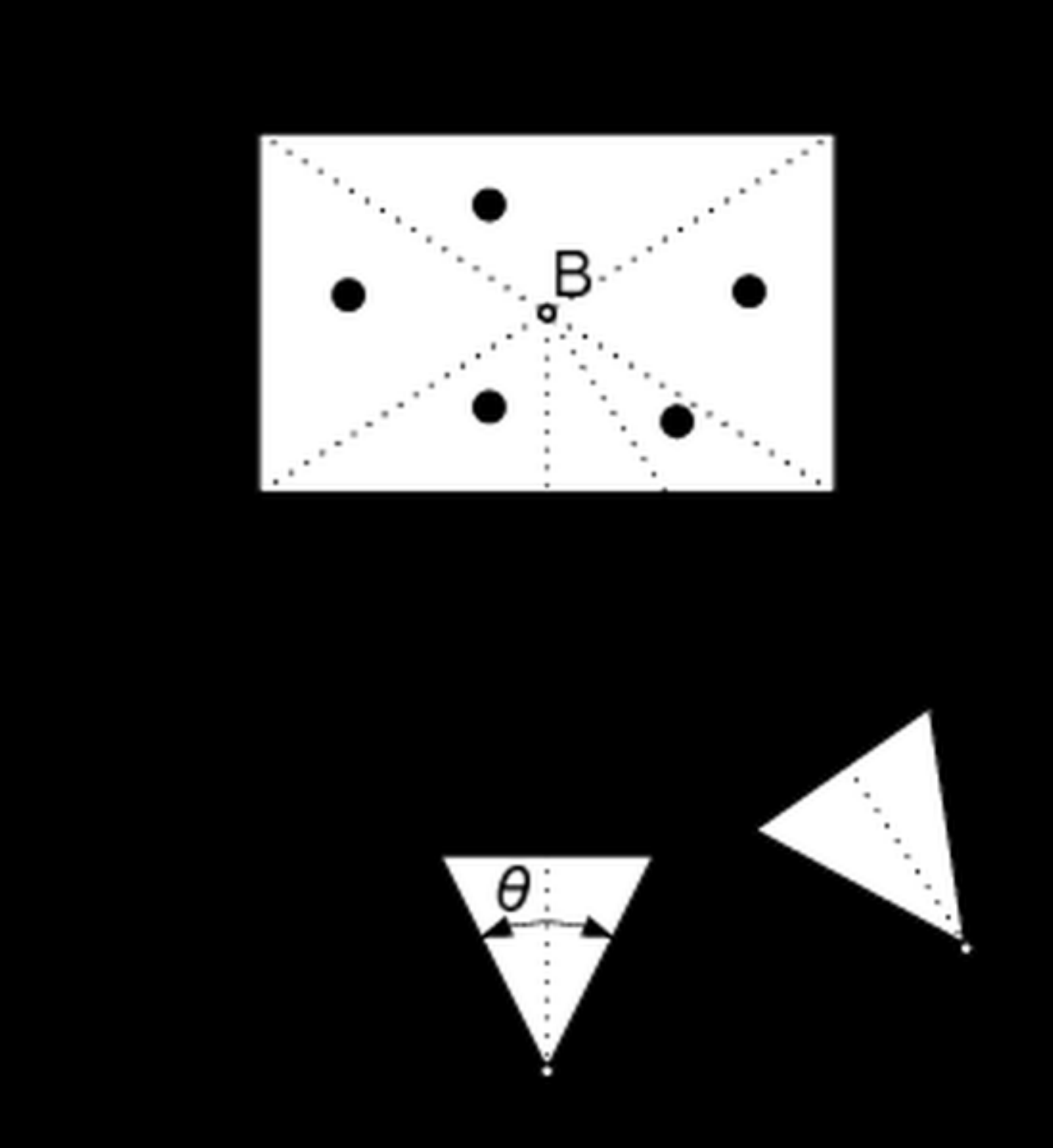}
\caption[]{Test environment used to generate synthetic data}
\label{jk:fig:4}
\end{figure}




Configuration used to generate synthetic data for experiments was similar to used in \cite{jk:nis04} and is depicted on Fig. \ref{jk:fig:4}.
Random 3D points are generated within an axis-aligned bounding box located in front of the camera. 
Camera focal local length $f=1$ and distance between front of the bounding box and camera optical center $dist = 4$.
Front of the bounding box subtends $\Theta = 45^{\circ}$ of visual angle.
Width of the bounding box is calculated from the formula:
$\frac{width/2}{dist} = \tan \Theta/2 \Rightarrow width = 2 dist \tan \Theta / 2$ and height is equal to width.
Height of the bounding box is equal to width.
It's assumed camera center $C_1$ is at the origin of world reference frame.
3D points are rotated by angle $\alpha$ around bounding box center $B$ and the rotation axis is parallel to the world coordinate frame $y$ axis. 
Suppose 3D point $X$ is rotated around a point $B$ by applying 3x3 rotation matrix $R$. The point coordinates in camera reference frame after rotation are $X' = R(X-B)+B = R(X-(B-R^{-1}B))$. This is equivalent to moving a camera to the new position $C_2$ by applying rotation $R$ and translating camera optical center by a vector $T = B-R^{-1}B$.

Generated 3D points are then projected onto image planes of both cameras and zero mean Gaussian noise is added to projection coordinates. 
In order to convert noise from pixel units to focal length units we assume x-resolution of a camera image plane is 1296 pixels (which corresponds to high resolution cameras). 
So $2 f \tan \Theta / 2 = 1296$ and 1 pixel corresponds to $1/648 f \tan \Theta /2$ of focal length units.

The aim of the experiment was to study accuracy of camera rotation and translation estimation based on noisy projection coordinates using our implementation of Nister 5-point algorithm and compare it to much simpler 7-point algorithm.
In each experiment $N=10000$ trials were performed. In each trial camera rotation $\hat{R}$ and translation $\hat{T}$ were estimated from noisy projection coordinates and compared with ground truth rotation $R$ and translation $T$.
Experiments were performed for 3 different configurations: almost-planar (depth to width ratio of the bounding box in which 3D points were generated = 0.01), semi-planar (depth to width ratio = 0.1), general (depth to width ratio = 1.0).

As metric reconstruction based on images from 2 calibrated cameras (only intrinsic parameters are known) is possible only up to a scale factor we cannot directly compare true translation vector $T$ and estimated translation vector $\hat{T}$.
Only angular component of translation error, that is an angle between true translation vector $T$ and estimated translation vector $\hat{T}$, is calculated using the formula:
$$T_{err} = \cos^{-1} \left( \frac{\hat{T}_i \cdot T}{| \hat{T}_i | |T| }\right)$$

Rotation error $R_{err}$ is measured as the rotation angle needed to align ground truth rotation matrix $R$ and estimated matrix $\hat{R}$.
$$
R_{err} = \cos^{-1} \frac{\mathrm{Tr}\left(\Delta R\right)-1}{2},
$$
where $\Delta R = R^{-1} \hat{R}$ is the rotation matrix that aligns estimated rotation $\hat{R}$ with the true solution $R$ and $\mathrm{Tr} ( \Delta R )$ is a trace of $\Delta R$.

\begin{figure}
\centering
\includegraphics[scale=0.44]{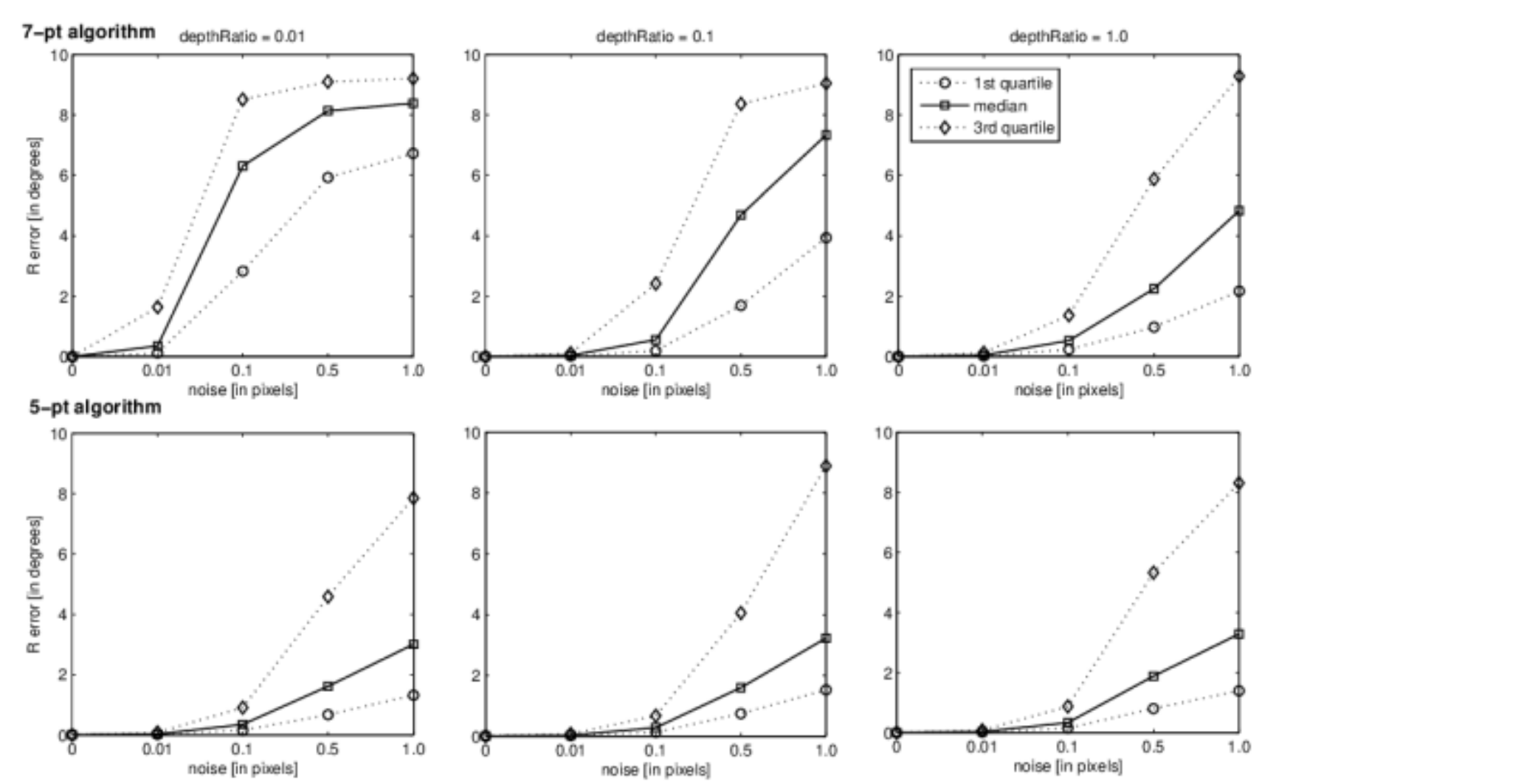}
\caption[]{Rotation matrix estimation error $R_{err}$ . Top row: 7-point algorithm, bottom row: 5-point algorithm. Columns from left to right correspond to almost-planar, semi-planar and general configurations.}
\label{jk:fig:20}
\end{figure}

\begin{figure}
\includegraphics[scale=0.44]{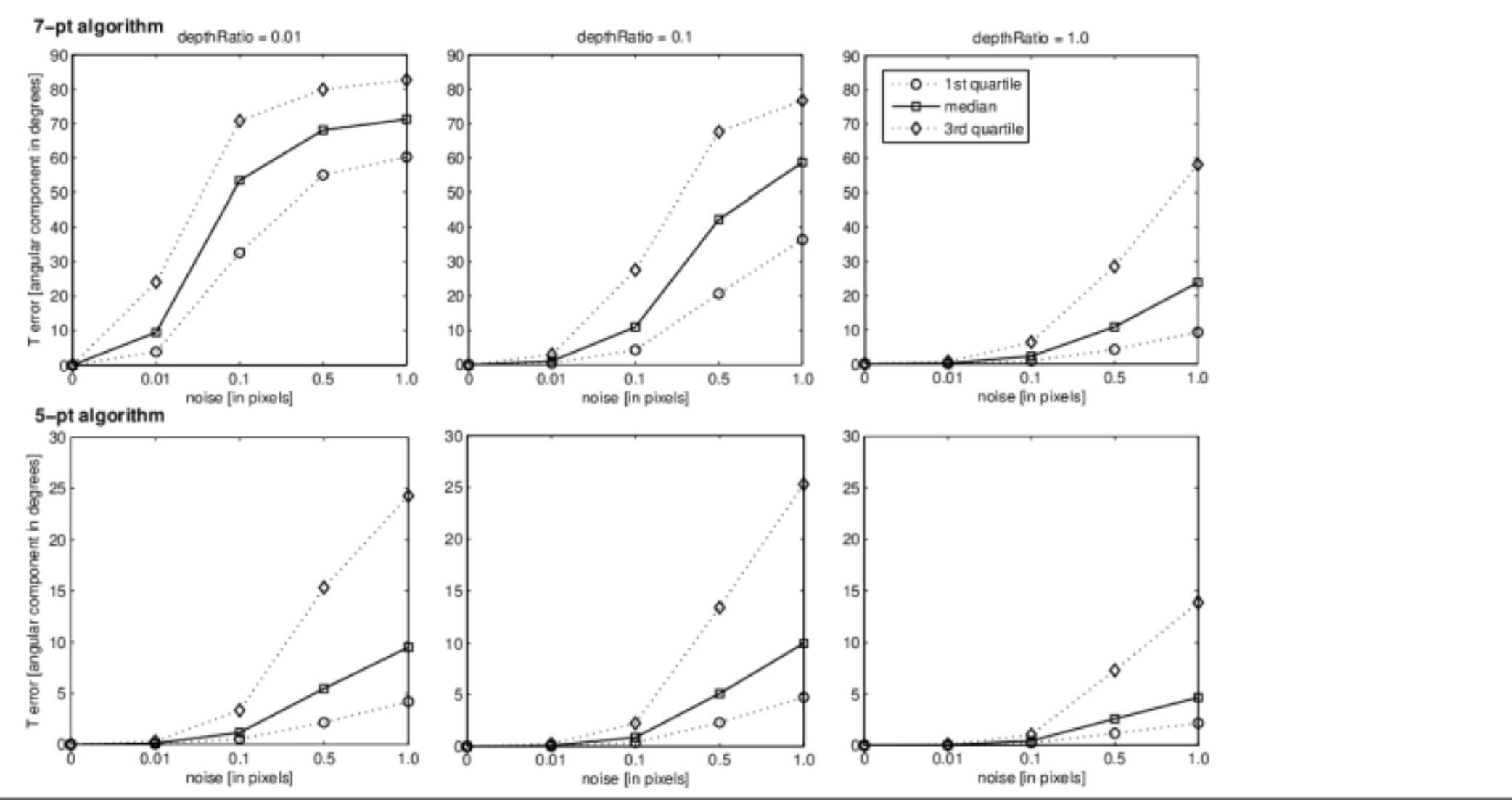}
\caption[]{Translation vector estimation error $T_{err}$ (angular component only). Top row: 7-point algorithm, bottom row: 5-point algorithm.
Columns from left to right correspond to almost-planar, semi-planar and general configurations.}
\label{jk:fig:21}
\end{figure}

Experiment results are depicted on Fig. \ref{jk:fig:20} and \ref{jk:fig:21}.
Fig. \ref{jk:fig:20} presents median, first and third quartile of rotation matrix estimation errors $R_{err}$ for $N=1000$ trials. It can be seen that for all tests scenarios: almost-planar (left column), semi-planar (middle column) and general (right column) rotation error $R_{err}$ median is lower for 5-point algorithm than for 7-point algorithm. Additionally 5-point algorithm does not suffer from planar degeneracy, it performs equally well in all 3 configurations (almost planar, semi planar and general). Performance of 7-point algorithm deteriorates significantly when 3D point configuration becomes more planar, which is in line with theoretical results.

Results for translation error $T_{err}$ are presented on Fig. \ref{jk:fig:21}.
Again 5-point algorithm outperforms 7-point algorithm in all 3 configurations. It can be noted that 5-point algorithm translation error increases when 3D point configuration becomes more planar but still is significantly lower than 5-point algorithm error.





\paragraph{Experiments on real world data}

\begin{figure}
\centering
\includegraphics[height=9 cm]{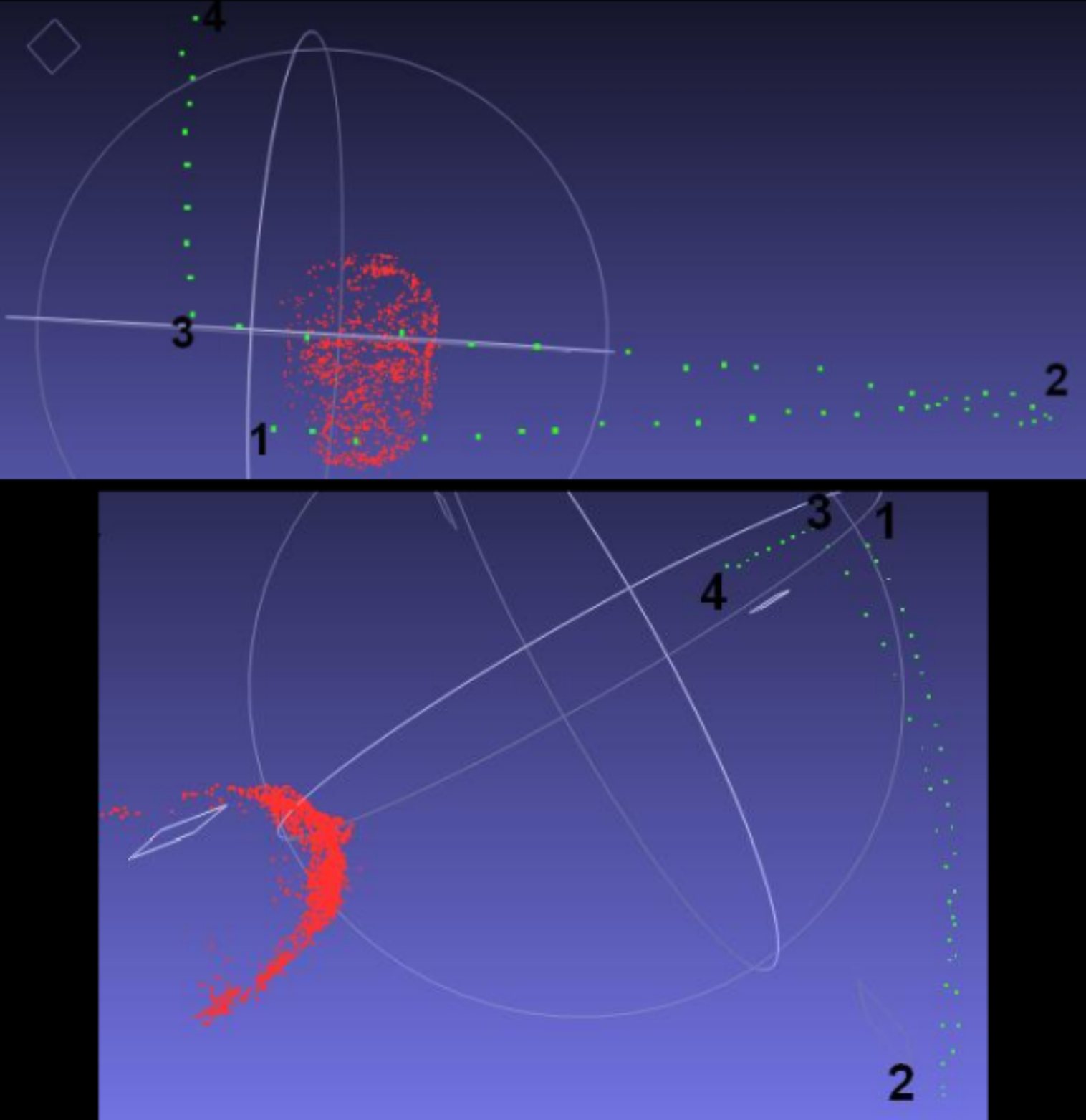}
\caption[]{Estimated camera positions (green dots) and 3D points (red dots) recovered from sequence depicted on Fig. \ref{jk:fig:10}. Head is initially positioned in front of the camera (1) and rotates right app. 45$^{\circ}$(2), back to the frontal pose (3) and downward app. 20$^{\circ}$ (4).}
\label{jk:fig:90}
\end{figure}

Camera pose reconstruction results on a real world data was verified using a number of input sequences containing images of a head rotating in front of the camera.
Camera pose estimation results on exemplary input sequence depicted on Fig. \ref{jk:fig:10} are shown on Fig. \ref{jk:fig:90}. Results are as expected: head is initially placed in front of the camera and it corresponds to camera placed at position 1, then as head rotates rightward camera moves along a circular trajectory up to position 2, then head rotates leftward, back to a frontal pose, and camera moves backward along a circular trajectory to position 3, finally head is rotates downward which corresponds to camera moving up to position 4.
But due to lack of a reliable ground truth data it was not possible to evaluate these results quantitatively.

\section{Conclusions and Future Work}
\label{jk:sec:4}

Making an assumption about camera calibration (known intrinsic parameters) allows to find a relative pose between 2 cameras using only 5 pairs of corresponding points instead of 7 pairs needed in uncalibrated setting. This can make a difference when processing images of a low textured objects where a number of reliable matches between features on 2 images is low.
It was also quantitatively verified using synthetic data that Nister solution to 5-point relative pose problem, despite its complexity, is numerically much more stable than 7-point algorithm. Additionally performance of 5-point algorithm doesn't deteriorate when 3D points configuration becomes more planar.

In the future it's planned to use the presented method as a first stage in a dense stereo reconstruction system.
After camera pose is estimated for each image in the sequence some multi-view stereo reconstruction method will be used to generate a dense point cloud representing an object.


\begin{thebibliography}{99.}
%
%
%


\bibitem{jk:fis81} Fischler M, Bolles R (1981)
Random Sample Consensus: A Paradigm for Model Fitting with Applications to Image Analysis and Automated Cartography. Communications of the ACM

\bibitem{jk:har04} Hartley R, Zisserman A (2004)
Multiple View Geometry in Computer Vision.
Cambridge University Press

\bibitem{jk:har94} Haralick R, Lee C, Ottenberg K, Nolle M (1994)
Review and analysis of solutions of the three point perspective pose estimation problem.
International Journal of Computer Vision

\bibitem{jk:nis04} Nister D (2004)
An efficient solution to the five-point relative pose problem.
IEEE Transactions on Pattern Analysis and Machine Intelligence

\bibitem{jk:lowe99} Lowe, D (1999)
Object recognition from local scale-invariant features.
Proceedings of the International Conference on Computer Vision

\bibitem{jk:snav07} Snavely, N et al. (2007)
Modeling the World from Internet Photo Collections
International Journal of Computer Vision

\bibitem{jk:tom91} Tomasi C, Kanade T (1991)
Detection and Tracking of Point Features. 
Carnegie Mellon University Technical Report

\bibitem{jk:tri99} Triggs B, McLauchlan P, Hartley R, Fitzgibbon A (1999)
Bundle Adjustment — A Modern Synthesis.
Proceedings of the IWVA

\bibitem{jk:sei06} Seitz, S. et al. (2006)
A Comparison and Evaluation of Multi-View Stereo Reconstruction Algorithms.
CVPR 2006, Vol. 1, 2006

\end{thebibliography}
\end{document}